\documentclass[runningheads]{llncs}
\usepackage[T1]{fontenc}
\usepackage{graphicx}
\usepackage{booktabs}
\usepackage[misc]{ifsym}

\usepackage{mwe}
\usepackage{multirow} 
\usepackage{amsmath}

\begin{document}

\title{Bias Amplification in Multi-Agent Network: How Biased Agents Shape Opinions and Rhetoric }

\author{Omran Berjawi\inst{1} \and
Giuseppe Fenza\inst{2} \and
Rida Khatoun\inst{1}}


\authorrunning{O. Berjawi et al.}

\institute{Institut Polytechnique de Paris, Télécom Paris, Palaiseau, France
\email{\{omran.berjawi,rida.khatoun\}@telecom-paris.fr}
\and
University of Salerno, Fisciano, Italy
\email{gfenza@unisa.it}}

\maketitle

\begin{abstract}
Large language models (LLMs) are increasingly deployed in applications involving interaction between agents, where their output plays a role in collective reasoning and decision-making processes. Despite significant research into the functioning of LLMs in such multi-agent systems, the processes of bias propagation in such systems are still a challenge. This work studies how biased opinions are propagated in the form of textual interaction in an environment of LLMs, in which a minority of agents maintain persistent extreme opinions, while the remaining agents iteratively update their beliefs through structured textual interactions. The findings show that even the presence of a small percentage of biased agents in such a system leads to significant shifts in the opinions of non-biased agents. It suggests that for the same percentage of biased agents, the shifts occur more quickly for the Llama~3.2 model when compared to a classical Friedkin-Johnsen (FJ) model.  Further semantic analysis demonstrates that rhetorical consistency in textual explanations increases systematically with biased exposure and, importantly, is partially decoupled from numerical convergenumericalutral agents adopt the vocabulary employed by the biased agents even in configurations where their numerical opinion shifts remain moderate. The research helps explain how bias and language develop together in multi-agent language model ecosystems.

\keywords{Bias Amplification \and Opinion Dynamics    \and Large Language Models (LLMs) \and  Multi-Agent Systems}
\end{abstract}

\section{Introduction}
The increasing deployment of large language model (LLM) agents as decision-making systems in most fields transforms how information is consumed and what narratives get formed within public discourse~\cite{gallegos2024bias} , much as rhetorical strategies of influential actors shape opinion in human social media~\cite{berjawi2025analyzing}. Yet LLMs are far from being neutral technologies, inheriting the biases inherent in the data used to train them as well as various societal assumptions. This makes the study of how these models propagate and intensify biases during interactions between them and humans crucial not only for responsible development but also for ensuring integrity in AI-mediated communication. It is well understood that the social dynamics of humans make opinions converge on either end of the spectrum. Social phenomena such as echo chambers, selective exposure, and ingroup bias contribute significantly to polarization among real-life communities~\cite{berjawi2024multi}. It has been found that a relatively small proportion of people biased towards one direction of thought can have a significant influence on the overall opinion within the community, driving it to even greater polarization in a non-linear and unpredictable manner. Bounded confidence model as well as iterative updating have proven effective in modeling such opinion dynamics~\cite{degroot1974reaching,hegselmann2015opinion}. However, such models use abstract numerical representations and cannot represent language-based interaction processes.

With the rise of LLMs in consequential tasks, including policy consultation, content moderation, or AI-assisted deliberation, an interesting issue arises as to whether bias tendencies in interactions among humans may be observed among the LLM agents due to their direct implications for algorithmic fairness in deployed AI systems. Some studies have investigated this issue from different perspectives, such as studying polarization and opinion formation among LLM agents~\cite{piao2025emergence}. ~\cite{borah-mihalcea-2024-towards} found that agents exhibiting implicit biases worsen over repeated interactions.  ~\cite{chuang-etal-2024-simulating} showed that prompt design can disrupt consensus formation within LLM networks. However, the effect on community opinion when neutral agents interact with biased agents over a long period is still a challenge. To address this issue, we study two hypotheses that were observed from earlier work: 

\begin{itemize}
    \item \textbf{H1:} \textit{Neutral agents who are systematically
    exposed to biased sources will demonstrate a significant shift of
    their initial positions toward the direction of the encountered
    bias.}
    \item \textbf{H2:} \textit{Neutral agents who engage in iterative interactions with biased agents will progressively converge in their linguistic patterns, with rhetorical alignment emerging even where numerical opinion shifts remain moderate.}
    
\end{itemize}

To investigate these two hypotheses, we are simulating a discussiregulations LLM agents wi a fully connected network. These agents discuss four controversial issues: AI safety regulation, vaccine mandates, immigration policy, and climate change. In these simulations, some agents hold extremely fixed views, while others adjust their opinions gradually with a limit on how much they can change. Our analysis looks at both the direction of opinion spreading and how the meaning in agents' textual justifications evolves over discussions. Our results indicate that increasing exposure to persistently biased sources causes neutral agents' opinions to shift, with more significant effects when the bias level is higher. Compared to a traditional FJ baseline, interactions using Llama~3.2 converge more quickly and veer more towards the extremes in terms of bias. Textual justifications also show consistent rhetorical alignment with these changes in opinion, covering all four topics we looked at. 

In summary, this paper makes three main contributions: first, a simulation framework using multi-agent LLMs to study how opinions evolve with controlled exposure to biased sources; second, a comparison between classic model and LLM interactions showing different behaviors in how quickly opinions settle; and third, proof that in networks of LLM agents, both opinions and the way language is used become more aligned with bias over time.

The rest of this paper follows this structure: Section~\ref{sec:Related_Work} talks about earlier work on opinion dynamics, bias amplification, and LLM-based agents. Section~\ref{sec:Methodology} describes the proposed framework. The setup and results of the experiments are in Sections~\ref{sec:Experiments} and~\ref{sec:Results}. Discussions on the results and limitations are in Sections~\ref{sec:Discussion} and~\ref{sec:Limitations}. Section~\ref{sec:Conclusion} concludes the paper.

\section{Related Work}
\label{sec:Related_Work}

The study of opinion dynamics and bias propagation has a long tradition in computational social science. Classic models (e.g., DeGroot's Averaging Model, Bounded Confidence) are being extended to examine the impact of both the topology of the underlying social network and various forms of "social influence" on group opinion formation ~\cite{degroot1974reaching,hegselmann2015opinion}. Simultaneously, empirical investigations of real-world social networks have found that selective exposure to information, the existence of echo-chambers, and in-group identity all contribute to polarization and the separation of groups into separate ideological communities ~\cite{bakshy2015exposure}. These two areas provide essential theoretical foundations for understanding how prejudice can develop and grow throughout the interactions of individuals.

As large language models have become increasingly prevalent in research, there is a growing interest among computational social scientists to apply them as tools for simulating a variety of social phenomena, including opinion development and polarization. In early experiments, it was shown that LLM-based interaction may create simulated polarized behavior among simulated humans in much the same way that actual humans do by creating 'echo chamber'-like clusters of individuals who share the same views and by displaying levels of homophily cluster formation similar to those observed in human social networks ~\cite{piao2025emergence}. Later studies have also developed multi-agent simulation frameworks using LLscales moddegreesplex social behaviors, including multi-topic echo chamber development in a single simulated population ~\cite{turn0academia20} and simulated social media event dynamics in which agents exhibit varied attributes ~\cite{turn0search2}. Additionally, other works have investigated how agent-specific properties, such as their composition of mindset types and community structures, affect changes in opinion among simulated humans in LLM-driven simulations; these results demonstrate that dominant mindsets are capable of shLLMsng the collective opinion environment ~\cite{turn0search0}. More recent proposals for hybrid models combine traditional equation-based dynamics with LLM-based agents to predict the futurLLMsrajectory of social opinion ~\cite{turn0search1}; still further explorations of social influence dynaLLMss have been conducted by analyzing conformity,LLMslarization, and fragmentation at varying model scale and degree of reasoning capability ~\cite{turn0academia26}.

Other works show that the biases increase when a large number of LLM agents interact over time.\cite{turn0search10} demonstrate that homophily can lead to biased structures in the agents' network. Evaluating LLMs against survey ground truth, \cite{turn0search6} further report that simulation accuracy varies substantially across countries and demographic groups, with better performance for Western, English-speaking populations. \cite{wang-etal-2025-decoding} compare LLM agents against classical models such as the FJ model, showing that LLMs can outperform classical models in simulating phenomena like echo chambers and polarization. 

In another line of research, \cite{cisneros-velarde-2025-biases} find that LLM reasoning affects group consensus when models are used for funding-allocation decisions. \cite{turn0search6} study the ability of LLMs to predict public opinion and find that they slightly improve accuracy over mathematical models owing to their richer reasoning capability.

Despite this progress, prior work leaves two gaps that we address here. First, existing multi-agent LLM studies examine polarization emerging from heterogeneous populations~\cite{piao2025emergence,chuang-etal-2024-simulating} or biases arising endogenously from model properties~\cite{borah-mihalcea-2024-towards,turn0search10}, but none isolate the causal effect of a persistently biased minority on an otherwise neutral population under controlled, graded levels of exposure. Second, existing studies track bias propagation exclusively through numerical opinion values, leaving the linguistic channel of influence unexamined. In contrast, our framework (i)~systematically varies the size of a fixed-stance biased minority while holding all other factors constant, (ii)~benchmarks the resulting dynamics against the classical FJ model under matched configurations, and (iii)~introduces a dual-layer analysis that jointly tracks numerical opinions and the semantic alignment of textual justifications. This dual-layer view reveals a partial decoupling between rhetorical and opinion convergence---rhetorical alignment reaches high levels even in configurations where numerical opinion shifts remain moderate---a phenomenon not previously documented in LLM multi-agent systems.
 
\section{Methodology}
\label{sec:Methodology}

\begin{figure}[t]
\centering
\includegraphics[width=0.5\textwidth]{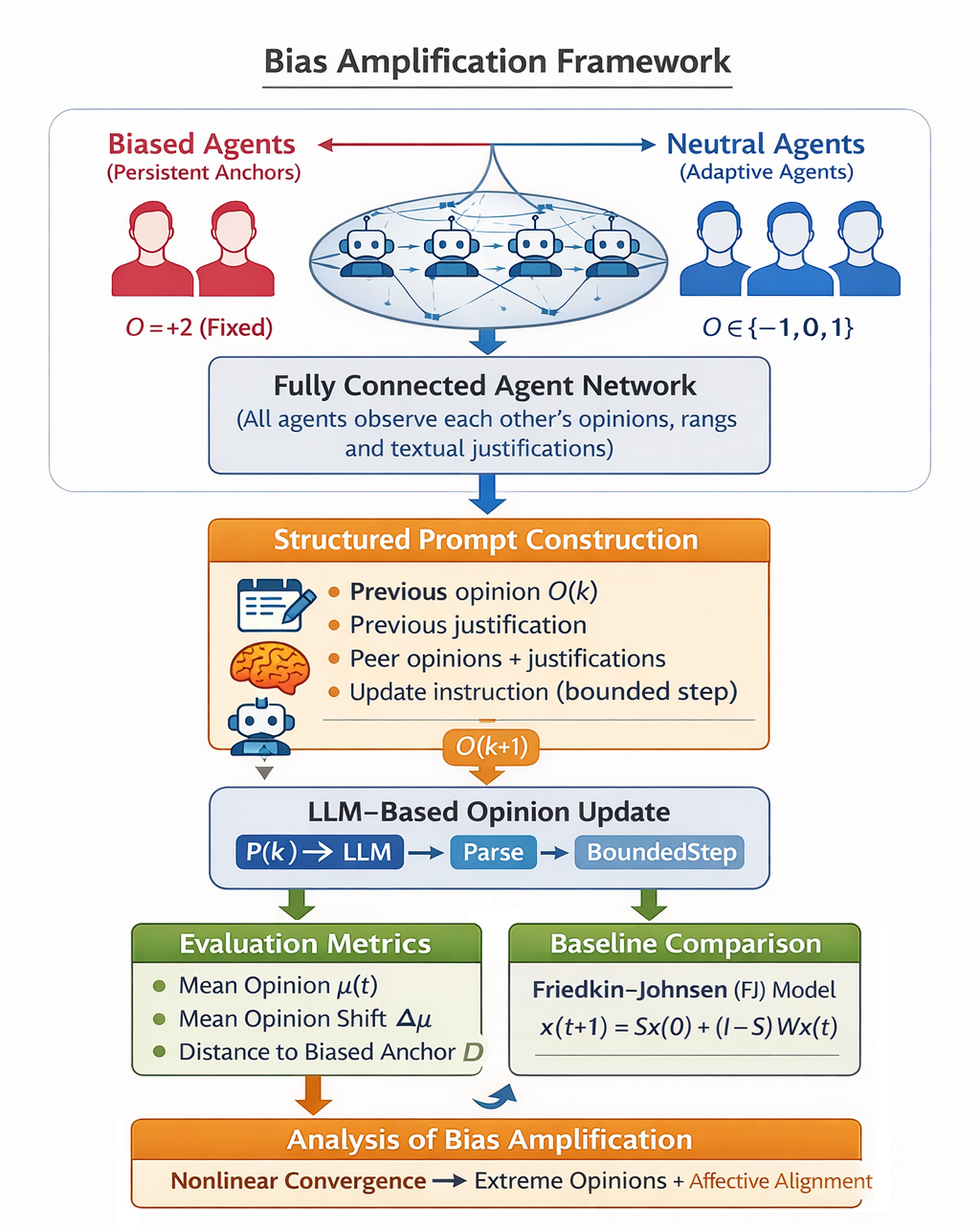}
\caption{Bias amplification Framework in LLM-driven agent networks.}
\label{fig:framework}
\end{figure}

This section presents the framework for simulating opinion dynamics in a network of LLM-driven agents. Figure~\ref{fig:framework} provides a high-level overview of the proposed framework, illustrating the interaction between agents and the opinion update process, and the comparison with a classical numerical baseline.

\subsection{Agent Network and Interaction Structure}
We consider a population of $N$ agents engaged in repeated interactions over a fixed network $G$. Each agent holds an internal opinion on a given topic, which evolves over a sequence of discussion rounds through exposure to the opinions and textual justifications of other agents. The agents are partitioned into two subsets: neutral agents, whose opinions evolve through interaction, and biased agents, which act as persistent agents by maintaining fixed extreme stances throughout the discussion. This design guarantees that each agent is exposed to the most extreme biased opinion at each round, enabling a direct test of hypotheses H1 and H2 (defined in Section~1). Formally, the network consists of agents $\{A_1, A_2, \dots, A_N\}$, where the neighbor set of each agent $A_i$ is given by: $\mathcal{N}_i = \{ j \mid j \in \{1, 2, \dots, N\}, \; j \neq i \}$.

\subsection{Agent Initialization}
Each agent \( A_i \) is modeled as an autonomous decision-making entity with a unique identifier. Every agent holds an initial opinion \( O_i \) on a topic \( T \in \{T_1, T_2, \dots, T_M\} \), represented by categorical numeric values \(\{-2,-1,0,1,2\}\). These values correspond to discrete stances: Strongly Disagree, Disagree, Neutral, Agree, Strongly Agree (see Table~\ref{tab:stance_values}). Extreme values (\(-2\) and \(+2\)) indicate maximal disagreement or agreement, while values near zero denote neutrality. In addition to numerical opinions, each agent generates a textual justification explaining its stance.  

Agents are initialized as either neutral (\( A_n \)) or biased (\( A_b \)). Neutral agents, forming the majority, are initialized with moderate opinions sampled uniformly from \(\{-1,0,1\}\). Biased agents are initialized with the extreme opinion value \(+2\). This one-sided bias enables controlled analysis of bias propagation. This directional design is intentional: by fixing the biased stance at a single extreme, we isolate the effect of sustained minority influence on neutral agents without conflating it with polarization dynamics that arise from competing biased factions. Note that although introducing $A_b$ agents at $+2$ shifts the mean of the \emph{total} population upward, all evaluation metrics (Section~4.5) are computed exclusively over neutral agents, and the mean opinion shift $\Delta\mu$ measures each neutral agent's movement relative to its own initial opinion. Any drift attributable to the initialization itself is therefore captured by the $A_b = 0$ baseline, against which all biased configurations are compared.

\begin{table}[t]
  \caption{Discrete opinion values and corresponding categorical stances.}
  \label{tab:stance_values}
  \centering
  \begin{tabular}{lc}
    \toprule
    \textbf{Opinion}   & \textbf{Value} \\
    \midrule
    Strongly Disagree  & $-2$ \\
    Disagree           & $-1$ \\
    Neutral            & $\phantom{-}0$ \\
    Agree              & $+1$ \\
    Strongly Agree     & $+2$ \\
    \bottomrule
  \end{tabular}
\end{table}
 
\subsection{Opinion Simulation}
Following initialization, agents iteratively update their opinions and textual justifications through interactions with all neighbors. At each discussion round \(k\), every agent \(A_i\) receives a structured prompt that provides the interaction context required for opinion updating.  The prompt is structured to separate prior beliefs, social input, and the decision-updating process. As illustrated in Figure~\ref{fig:prompt_template}, the prompt includes the agent's previous opinion $O_i^{(k)}$ and textual justification, the current opinions and justifications of all other agents in the network, and an instruction to update the opinion by at most one categorical step. Given this information, the LLM generates a response containing an updated categorical opinion together with a short justification. The resulting opinion update process can be summarized by the following transformation pipeline:

\[
P_i^{(k)}
\rightarrow
LLM(P_i^{(k)})
\rightarrow
\mathrm{Parse}(\cdot)
\rightarrow
\mathrm{BoundedStep}(\cdot)
\rightarrow
O_i^{(k+1)}
\]

\begin{figure}[t]
\centering
\small
\fbox{
\begin{minipage}{0.95\linewidth}
\setlength{\baselineskip}{0.7\baselineskip}

\textbf{Generalized Prompt Template for Agent $A_i$ at Round $k$}

\vspace{0.5em}

\textbf{Topic:} [Discussion Topic]

\vspace{0.5em}

\textbf{Your Previous Opinion:}  
Categorical stance: [Strongly Disagree / Disagree / Neutral / Agree / Strongly Agree]  
Numeric value: $O_i^{(k)}$

\vspace{0.5em}

\textbf{Your Previous Justification:}  
[Text explaining the reasoning behind the previous opinion]

\vspace{0.5em}

\textbf{Opinions from Other Agents:}

Agent $A_1$: stance + justification \\
Agent $A_2$: stance + justification \\
$\dots$

\vspace{0.5em}

\textbf{Instruction:}

Reflect on the discussion above. You may update your opinion based on the arguments presented.  
Your new opinion must differ from the previous one by at most one category step.

\vspace{0.5em}

Provide:

1. Updated categorical opinion  
2. Corresponding numeric value  
3. Short justification
\end{minipage}
}
\caption{Prompt used in each discussion round.}
\label{fig:prompt_template}
\end{figure}

First, the LLM processes the prompt \(P_i^{(k)}\) and produces a textual response \(LLM(P_i^{(k)})\), which includes an updated categorical stance (e.g., \textit{Agree}) and a brief justification. The parsing function \(\mathrm{Parse}(\cdot)\) then extracts the expressed stance and maps it to the corresponding numerical value in the discrete opinion set \(\{-2,-1,0,1,2\}\). Finally, the bounded-step operator ensures that the opinion can change by at most one category between successive discussion rounds. All opinions and justifications are recorded in JSON format for analysis.

\subsubsection{Opinion Update Mechanism}
The opinion update mechanism differs for neutral and biased agents. 

\begin{itemize}
    \item Neutral Agents (\( A_n \)) :  \( A_n \) update their opinions during each round $k$ according to:
\[
O_n^{(k+1)} = \mathrm{BoundedStep} \big( \mathrm{Parse}(LLM(P_n^{(k)})) \big),
\]

where $LLM(P_n^{(k)})$ produces the textual output including the updated stance and justification, $\mathrm{Parse}(\cdot)$ extracts the numerical opinion, and $\mathrm{BoundedStep}(\cdot)$ ensures at most a one-category change per round.

 \item Biased agents (\( A_b \)): \( A_b \) maintain a fixed numerical opinion throughout the simulation:
\[
O_b^{(k+1)} = O_b^{(k)}, \quad \forall k
\]

Although their numerical opinions remain fixed at \(+2\), biased agents regenerate textual justifications at each round, allowing them to adapt their rhetoric.

\end{itemize}

\section{Experimental Setup}
\label{sec:Experiments}
To evaluate the framework, we compare LLM-based simulations against a classical numerical FJ model. Agent counts, network structure, and bias configurations are kept identical across both conditions, with the opinion update mechanism as the only variable. This allows us to attribute any observed differences in dynamics directly to the nature of the interaction model rather than to structural differences in the experimental setup.

\subsection{Network and Simulation Parameters}
We run experiments on four discussion topics chosen to represent distinct policy domains to assess whether the amplification effects we observe are specific to one area or more general.

\begin{itemize}

  \item AI Safety Regulation ($T_{Reg}$): This topic centers on the tension between enabling technological progress and imposing precautionary constraints on AI developmeThe biased stance in our simulations favors stronger regulatory control.

  \item Vaccine Mandates ($T_{Vac}$): Here, agents discuss the trade-off between individual autonomy and collective health obligations, in which the biased agents support the vaccination mandates. 
  
  \item Immigration Policy ($T_{Imm}$): The new regulations to reduce the immigration phenomenon are used as a topic of discussion. The biased agents agree with stricter policies.
  
  \item Climate Change ($T_{CLIM}$): This topic discusses the concerns regarding climate change, in which biased agents support the aggressive intervention to reduce this phenomenon.
  
 \end{itemize}

All simulations use a population of $N = 50$ agents interacting over a fully connected network. For each of the above topics, simulations are run under varying bias levels $A_b \in \{0, 2, 4, 6, 8, 10\}$, where $A_b$ denotes the number of biased agents, corresponding to biased minorities ranging from $0\%$ to $20\%$ of the population. The configuration $A_b = 0$ serves as a fully neutral baseline containing no biased agents, while increasing $A_b$ introduces progressively stronger minority influence.

\subsection{Llama 3.2 Implementation}
LLM agents are implemented based on the Llama 3.2-8B model using the LangChain framework, in which each agent is considered an independent LangChain agent. The interaction between agents is performed using a structured prompt, as detailed in the methodology (see Section~\ref{sec:Methodology}). All agents are configured with identical parameters across discussion rounds, using a temperature of $0.7$.

\subsection{Textual Evaluation}
To complement the analysis of numeric opinion evolution, we examine the affective content of agents’ textual justifications across discussion rounds. We quantify the semantic alignment of neutral agents’ justifications relative to biased agents to capture whether neutral agents increasingly adopt the rhetorical tone of biased agents oCosine, even if their numeric opinions remain moderate.

For each agent, the textual justification generated at each round is encoded into sentence emwithngs using a pre-trained SBERT model. A cosine similarity is computed between each neutral agent’s text and the set of biased agent justifications in the same round. The mean alignment score at round \(k\) is given by:
\[
A^{(k)} = \frac{1}{N_n} \sum_{i=1}^{N_n} \overline{\mathrm{sim}}\Big(J_i^{(k)}, \{ J_j^{(k)} \}_{j \in \mathcal{A}_b}\Big),
\]

where \(A^{(k)}\) is the average semantic alignment at discussion round \(k\), \(J_i^{(k)}\) and \(J_j^{(k)}\) are the textual justifications of neutral agent \(A_i\) and biased agent \(A_j\), respectively, and \(\overline{\mathrm{sim}}(\cdot)\) denotes the mean cosine similarity between a neutral agent’s justification and all biased agents’ justifications.

 \begin{figure}[t]
\centering
\includegraphics[width=\textwidth]{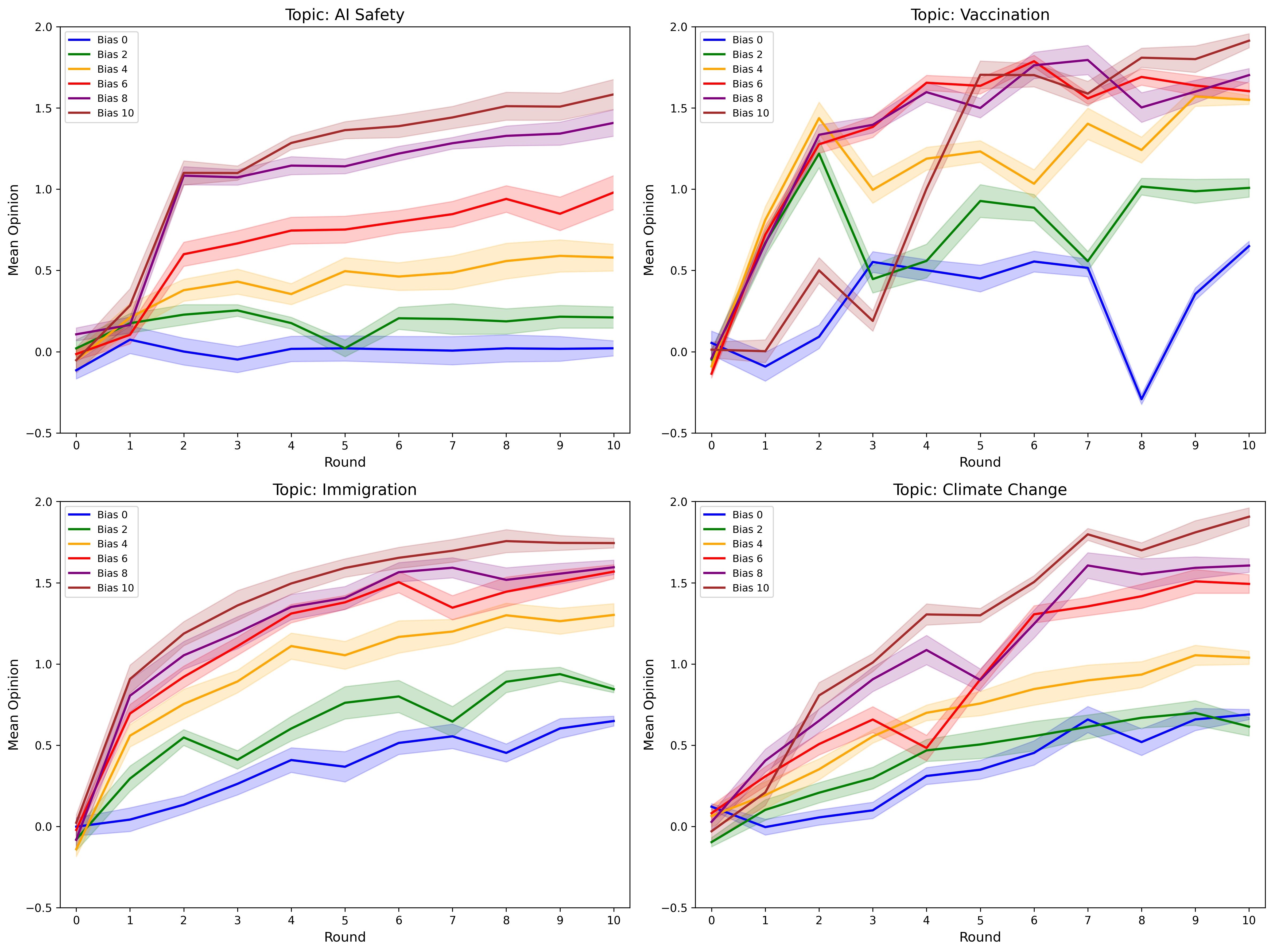}
\caption{Evolution of mean opinion trajectories across bias levels in the Llama~3.2 model. Shaded bands denote 95\% confidence intervals computed from the standard error across various runs.}
\label{fig:LLMdistirubtion}
\end{figure}

\begin{table*}[t]
\centering
\caption{The Bias metrics of models across four topics.}
\label{tab:fj_llm_topics_comparison}
\resizebox{\textwidth}{!}{ 
\begin{tabular}{c|cc|cc|cc|cc|cc}
\hline
\multirow{2}{*}{\textbf{Bias Level}}
& \multicolumn{2}{c|}{\textbf{FJ Model}}
& \multicolumn{2}{c|}{\textbf{Llama 3.2 (AI Safety)}}
& \multicolumn{2}{c|}{\textbf{Llama 3.2 (Vaccine Mandates)}}
& \multicolumn{2}{c|}{\textbf{Llama 3.2 (Immigration)}}
& \multicolumn{2}{c}{\textbf{Llama 3.2 (Climate Change)}} \\
& Mean Shift & Dist. to Anchor
& Mean Shift & Dist. to Anchor
& Mean Shift & Dist. to Anchor
& Mean Shift & Dist. to Anchor
& Mean Shift & Dist. to Anchor \\
\hline
0  & -0.05 & 2.05 & 0.02 & 1.97 & 0.66 & 1.34 & 0.78 & 1.38 & 0.77 & 1.31 \\
2  & 0.10 & 1.90 & 0.25 & 1.74 & 1.01 & 1.12 & 1.15 & 1.25 & 0.61 & 1.39 \\
4  & 0.25 & 1.75 & 0.68 & 1.31 & 1.55 & 0.45 & 1.51 & 0.54 & 1.05 & 0.95 \\
6  & 0.37 & 1.63 & 1.20 & 0.79 & 1.60 & 0.40 & 1.88 & 0.25 & 1.51 & 0.52 \\
8  & 0.43 & 1.57 & 1.46 & 0.53 & 1.72 & 0.30 & 1.79 & 0.38 & 1.62 & 0.41 \\
10 & 0.47 & 1.53 & 1.49 & 0.50 & 1.89 & 0.10 & 1.91 & 0.14 & 1.92 & 0.08 \\
\hline
\end{tabular}%
}
\end{table*}

\subsection{Baseline Comparison}
To contextualize opinion dynamics, we compare Llama 3.2 with the classical FJ model as a numerical baseline. The FJ model captures the evolution of opinions in a network by combining social influence from neighbors with agents’ intrinsic initial beliefs. For a population of \(N\) agents, the opinion of agent \(i\) at round \(k+1\) is updated according to:

\begin{equation}
O_i^{(k+1)} = \lambda_i O_i^{(0)} + (1 - \lambda_i) \sum_{j \in \mathcal{N}_i} w_{ij} O_j^{(k)}
\end{equation}

where \(O_i^{(0)}\) is the agent's initial opinion, \(\lambda_i \in [0,1]\) represents the agent’s susceptibility to social influence (self-weight), \(\mathcal{N}_i\) denotes the set of neighbors of agent \(i\), and \(w_{ij}\) are normalized weights representing the influence of neighbor \(j\) on agent \(i\) such that \(\sum_{j \in \mathcal{N}_i} w_{ij} = 1\). 

In our experiments, the FJ model is applied to the same configuration used for Llama~3.2: identical population size, network structure, and bias levels, with neutral agents' initial opinions sampled from $\{-1, 0, 1\}$ and biased agents anchored at $+2$. To ensure that the stubbornness assumptions are matched across the two conditions, biased agents are assigned $\lambda_b = 1$, making them fully stubborn and therefore exactly equivalent to the fixed-opinion biased agents in the LLM condition ($O^{(k+1)}_b = O^{(k)}_b = +2$ for all $k$), while neutral agents are assigned $\lambda_n = 0.5$, representing an equal balance between individual conviction and social influence. Each simulation is run for \(K = 10\) rounds. Under this configuration, the only factor that differs between the two conditions is the opinion update mechanism of neutral agents (numerical averaging versus language-mediated reasoning). So any observed differences in dynamics are attributable to the interaction mechanism itself rather than to asymmetric stubbornness assumptions. Each simulation is run for $K = 10$ rounds.

\subsection{Evaluation Metrics}
We quantify bias amplification using metrics computed exclusively over neutral agents:

\begin{itemize}

\item Mean Opinion: This metric measures the overall directional stance of the agents. 
\begin{equation}
\mu^{(t)} = \frac{1}{N_n} \sum_{i \in \mathcal{A}_n} o_i^{(t)} .
\end{equation}

\item Mean Opinion Shift: This metric captures the net change in opinions resulting from the interaction process. It measures the average movement of neutral agents between the initial and final discussion rounds:

\begin{equation}
\Delta \mu =
\frac{1}{N_n}
\sum_{i \in \mathcal{A}_n}
\left( o_i^{(T)} - o_i^{(0)} \right).
\end{equation}

\item Distance to Biased Anchor: This metric measures the average absolute distance between the final opinions of neutral agents and the biased anchor value, capturing the degree of convergence toward the imposed extreme stance. 
\begin{equation}
D =
\frac{1}{N_n}
\sum_{i \in \mathcal{A}_n}
\left| o_i^{(T)} - o_{\text{bias}} \right|.
\end{equation}

\end{itemize}

\subsection{Reproducibility and Statistical Robustness}
Because LLM outputs are stochastic even under fixed prompts, we run each simulation configuration 15 times independently. For the FJ model, the dynamics are deterministic given fixed agent opinions; we nevertheless run 15 independent trials with different random initializations to keep the comparison fair. All findings illustrated in the figures use 95\% confidence intervals calculated from standard errors across runs.

All results are reported as means with 95\% confidence intervals, estimated from standard errors across runs and displayed as shaded bands around trajectory curves in the figures.

\section{Experimental Results}
\label{sec:Results}

\subsection{Opinion Dynamics and Bias Amplification}
\subsubsection{Temporal Evolution of Opinion Trajectories}
\begin{figure}[!t]
    \centering
    \includegraphics[width=0.8\columnwidth]{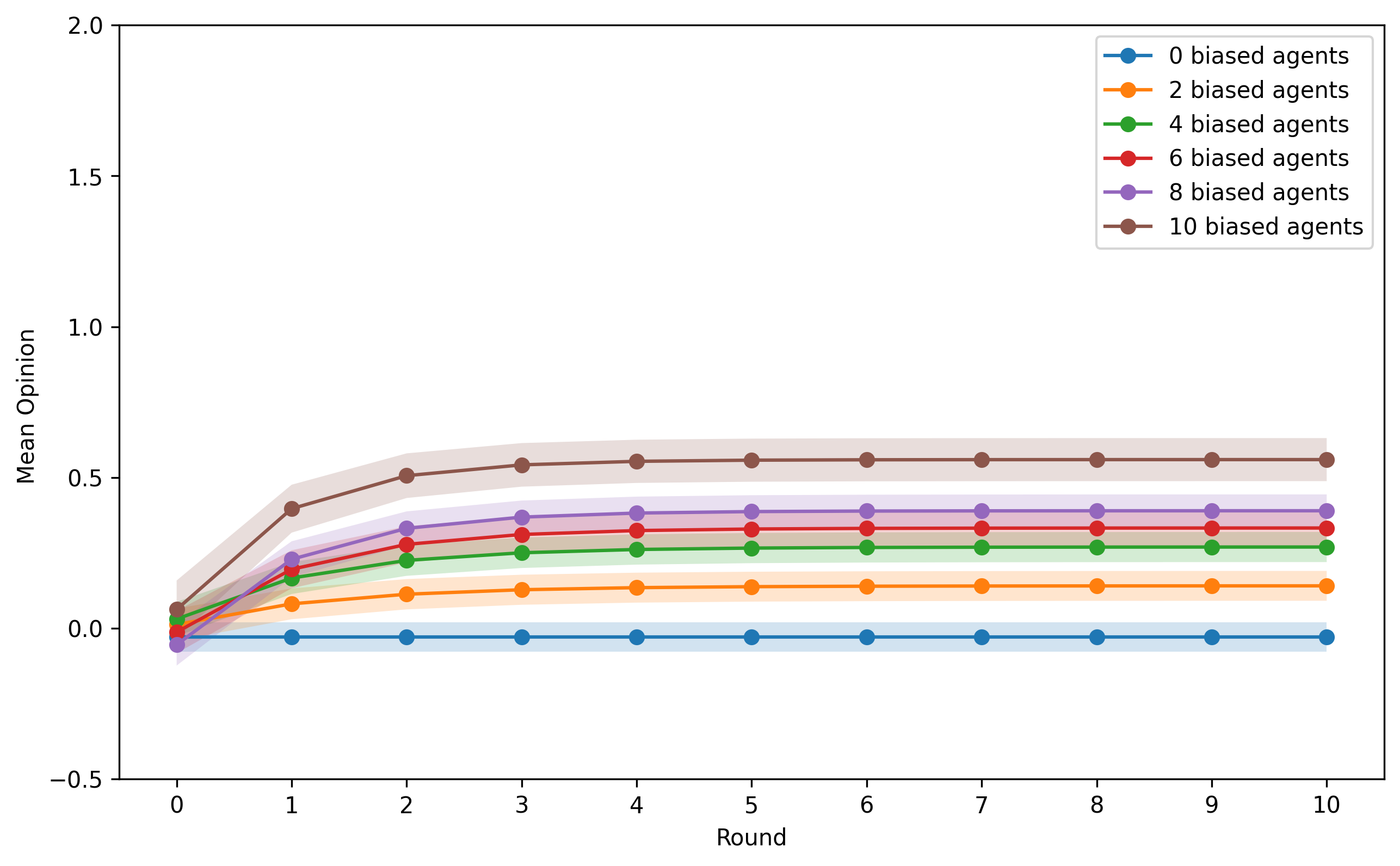}
    \caption{Evolution of opinion distributions across bias levels for FJ Model}
    \label{fig:distribution}
\end{figure}

. Figures~\ref{fig:LLMdistirubtion} and~\ref{fig:distribution} show the evolution of mean opinion for the Llama~3.2 and FJ models across the four topics. 

We can see that the Llama 3.2 curves show an increase in the mean opinion when the number of $A_b$ increases. Starting with the baseline configuration when ($A_b=0$), the average opinion remains close to the neutral values over the simulation wi,th a little bit of a shift over time. When the $A_b$ increases, the opinion shifts upward, in which the mean opinion becomes larger when the configurations are at a high level ($A_b=8,10$). This pattern is found across all topics \(T_{Reg}\), \(T_{Vac}\), \(T_{Imm}\), and \(T_{CLIM}\), although the magnitude of increase varies modestly across topics. It is important to note that most of the opinion change occurs within the first few rounds ($k \leq 3$), after which the trajectories gradually stabilize. 

We also observe a transient dip in the unbiased configuration ($A_b = 0$) for $T_{Vac}$ around round~8, reflecting temporary downward revisions by several neutral agents rather than a measurement artifact. Such oscillations are expected in the absence of a persistent anchor, where collective opinion is more sensitive to the arguments generated at each round.

On the other hand, the findings for the FJ model (Figure~\ref{fig:distribution}) demonstrate gradual transitions over time. For all configurations, the increase in mean opinion follows a linear progression over rounds, with a minimal difference between bias levels.

\subsubsection{Aggregate Bias Metrics Across Topics}
Table~\ref{tab:fj_llm_topics_comparison} shows the aggregate bias metrics for each model across topics, combining avthe erage shifts in stance with movement toward preset biases. While one model pulls closer to fixed viewpoints, the other spreads more evenly, both measured by deviation and directional tilt. 

Starting from $A_b=0$, the FJ model shows a slow climb in average opinion shift, reaching nearly half a point when bias at $A_b=10$. As that happens, the gap to the skewed reference tightens, dropping just $2.05$ to $1.53$. Step by step, each rise in bias nudges opinions slightly closer. Not dramatic, just a steady lean toward the pull is of influence.

One thing stands out in the Llama 3.2 setup: opinions shift considerably more, regardless of topic. Take AI Safety: when $A_b=0$, the average shift is just $0.02$, but it reaches $1.49$ at $A_b=10$. Over the same range, the distance to the anchor drops from $1.97$ to $0.50$. The same pattern is observed across \(T_{Vac}\), \(T_{Imm}\), and \(T_{CLIM}\). Each one peaks in shift size under the strongest bias configuration, approaching the biased anchor most closely. Though numbers change depending on the subject, the overall pattern remains consistent.

A key observation is that vaccine mandates show a clear shift in opinion even when bias is set to zero ($A_b = 0$, mean shift $= 0.66$). With no push from biased individuals, neutral ones still drift toward support. This represents a model-level bias, producing bigger effects when biased agents are around. It shows that language-based interactions amplify the impact of minorities more than classical models suggest.

\subsection{Affective Alignment Analysis}
\begin{figure}[!t]
    \centering
    \includegraphics[width=0.8\columnwidth]{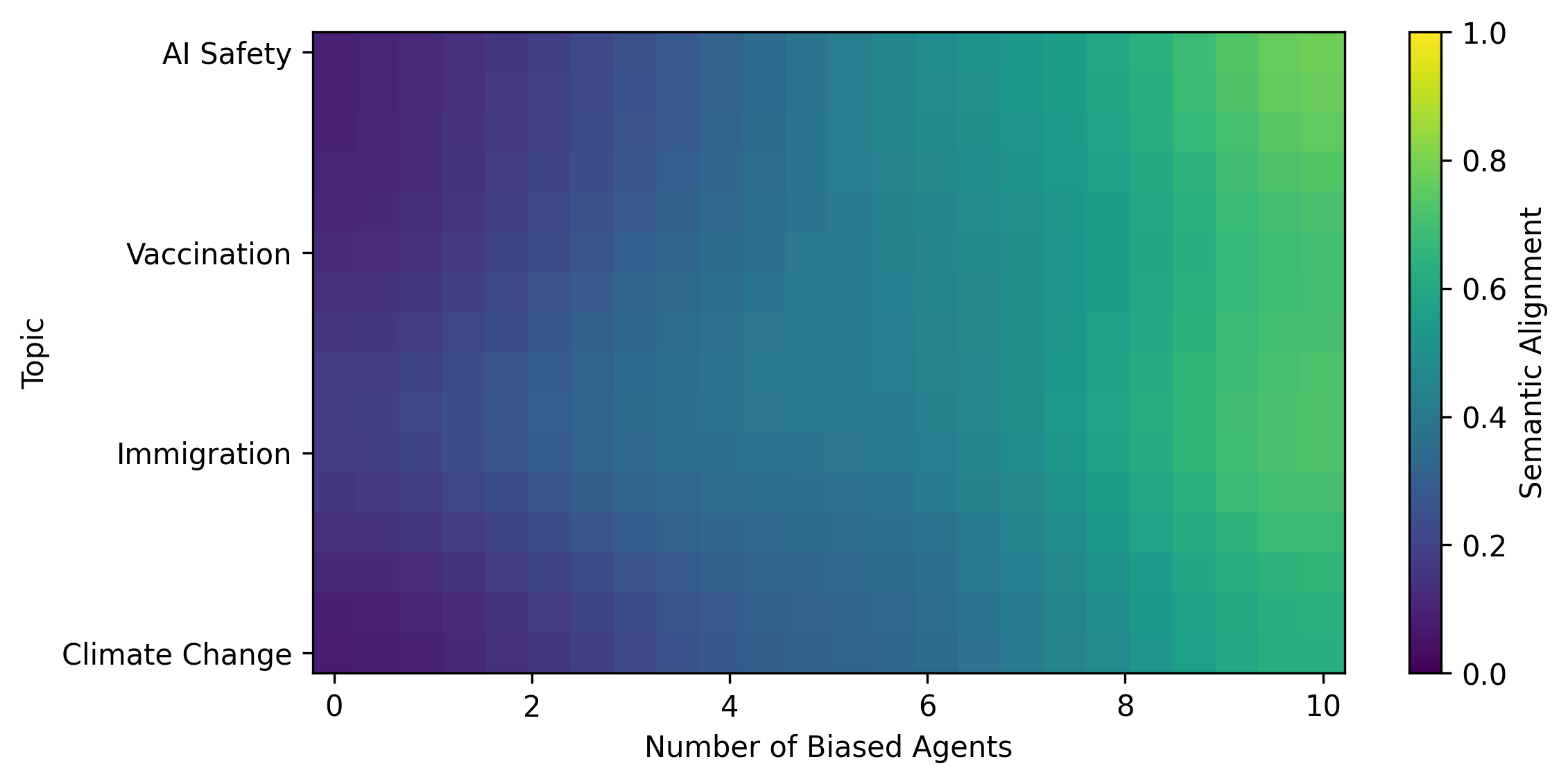}
    \caption{Rhetorical alignment of neutral agents across bias levels in the Llama 3.2 model.}
    \label{fig:Semantic}
\end{figure}

Figure~\ref{fig:Semantic} presents the semantic alignment of textual justifications generated by neutral agents across topics and bias configurations. For each configuration, alignment is computed per neutral agent as the mean cosine similarity between its justification and those of all biased agents at the final discussion round (Section~4.3), then averaged over all neutral agents and over the 15 independent runs. This yields one alignment value per (topic, bias level) pair, i.e., $4 \times 6 = 24$ measured values in total, which form the cells of the heatmap.


Across all topics, semantic alignment increases monotonically with the number of biased agents. For the baseline configuration ($A_b=0$), alignment values remain low, indicating limited similarity in textual justifications among agents. As the number of biased agents increases, alignment values rise steadily, with the highest levels observed at $A_b=10$.

The rate of increase differs across topics. \(T_{Reg}\) exhibits the highest alignment values at larger bias configurations, followed closely by \(T_{Vac}\) and \(T_{Imm}\). \(T_{CLIM}\) shows a similar upward trend but with comparatively lower alignment values across all configurations. Despite these differences in magnitude, the overall pattern of increasing alignment with higher biased agent proportions is consistent across all domains.

Note that both axes of the heatmap are discrete: alignment is measured only at the six bias configurations and four topics, and the smooth appearance of the figure results from bilinear interpolation applied for visual readability. The measured values themselves show a gradual transition from low to high alignment as bias increases, with no abrupt discontinuities. Intermediate configurations ($A_b = 4, 6$) display moderate alignment levels, forming a smooth progression between the neutral baseline and high-bias settings. This pattern indicates a continuous relationship between biased agent proportion and semantic convergence in generated justifications.

Crucially, comparing the alignment values in Figure~\ref{fig:Semantic} with the opinion trajectories in Figure~\ref{fig:LLMdistirubtion} reveals an asymmetry in convergence rates: semantic alignment begins rising substantially at intermediate bias configurations ($A_b = 4, 6$) even in topics where numerical opinion shift remains moderate. This temporal and quantitative decoupling, whereby language homogenizes faster and more completely than discrete opinions — constitutes the central empirical support for H2. It implies that an external observer monitoring only the textual output of such a system would overestimate the degree of opinion consensus among agents, since the rhetorical surface has converged while meaningful variation in underlying stances persists. This property is particularly relevant for fairness auditing in deployed systems, where semantic content rather than internal belief representations is typically the only observable signal.

\section{Discussion}
\label{sec:Discussion}
The results support both hypotheses and, taken together, point toward some broader implications for how multi-agent LLM systems should be monitored and audited.

\medskip
\noindent\textbf{H1:} \textit{Exposure to a minority of persistently biased agents leads to systematic opinion shifts in initially neutral agents.}

The observed opinion trajectories demonstrate that even at the lowest bias level we tested ($A_b = 2$), neutral agents show measurable directional drift, and this effect grows as more biased agents are added. The increase in mean opinion and the shrinking distance to the biased anchor both point in the same direction. This is consistent with Moscovici's minority influence account~\cite{moscovici1969influence}, which holds that a small, consistent minority can exert influence well out of proportion to its size. What makes the LLM case interesting is the amplification: Llama~3.2 produces faster and larger shifts than the FJ baseline, which suggests that language-mediated interaction, perhaps through the richness of the rhetorical context agents provide one another, This makes the system more susceptible to minority influence than classical numerical models predict.

\medskip
\noindent\textbf{H2:} \textit{Rhetorical alignment emerges alongside opinion shifts, partially masking cognitive diversity.}

The semantic alignment finding shows that as the proportion of biased agents grows, neutral agents' justifications increasingly resemble those of the biased minority. This holds smoothly across all topics, and alignment reaches high levels even in configurations where numerical opinions have shifted only moderately: agents adopt the rhetoric of the biased minority to a greater extent than they adopt its position. The result is a situation where, from the outside, a system can look more consensual than it actually is: language has homogenized while meaningful disagreement at the opinion level persists. This is not a subtle or marginal effect; it holds consistently across topics and bias levels. Whether rhetorical adoption also temporally precedes opinion change at the individual-agent level is a natural follow-up question requiring round-by-round alignment tracking.

\section{Limitations}
\label{sec:Limitations}
Several limitations should be considered when interpreting these 
findings. 

Single model and framework: All experiments were conducted with Llama~3.2-8B, orchestrated through the LangChain framework. Whether the decoupling between rhetorical and numerical convergence that we identify in H2 would appear with a different model, one with a different training distribution, scale, or alignment approach, is unknown at this point. Similarly, the results may be sensitive to the agentic framework used to structure interactions. We consider replication across model families (e.g., Mistral, Qwen, GPT-class models) and alternative agentic frameworks the most important open question raised by this work and a priority for follow-on research.

Prompt sensitivity: The bounded-step instruction and the overall framing of agent roles may be shaping responses in ways we have not fully disentangled from the bias effects of interest. We did not run systematic prompt ablations, and future work should do so to establish how robust the reportseveralare to variations in how agents are instructed.

Final-round alignment measurement: Semantic alignment was measured at the final discussion round. This establishes the co-occurrence of rhetorical and opinion convergence, and their difference in magnitude, but does not directly test temporal precedence. Verifying whether rhetorical alignment predicts subsequent opinion change at the individual-agent level requires per-round alignment tracking and lagged analyses, which we leave to future work.

One-sided bias: Biased agents were anchored exclusively at the positive extreme ($+2$). Since neutral agents drift toward support on some topics even in the absence of biased agents (Section~\ref{sec:Results}), indicating an intrinsic model-level prior, the amplification we measure may partially interact with this prior. Repeating the experiments with biased agents anchored at the opposite extreme ($-2$) would disentangle the effect of the promoted stance from that of the underlying model bias, revealing whether minority influence is symmetric or whether opinions are easier to shift in the direction of the model's prior than against it. We regard this as a natural next experiment within the present framework.

Taken together, these points position the paper as a proof-of-concept, one that establishes the measurability and direction of minority bias amplification in LLM agent networks, but that leaves a number of important questions open for future research.

\section{Conclusion}
\label{sec:Conclusion}
This paper examined bias amplification in multi-agent LLM networks by analyzing how persistent biased agents influence both opinion dynamics and semantic alignment. Through controlled simulations across multiple topics, we showed that increasing exposure to biased agents leads to systematic and nonlinear opinion shifts among neutral agents. Compared to the FJ baseline, Llama 3.2 interactions produce stronger and faster convergence toward biased positions. Beyond numerical opinion changes, our analysis revealed that semantic alignment in textual justifications increases consistently with biased exposure. This indicates that language convergence emerges alongside opinion shifts, even when some diversity in discrete opinions remains.  These findings highlight the dual role of LLM interactions in shaping both beliefs and their expression, raising important considerations for the deployment of AI systems in socially sensitive contexts. Future work can extend this framework to heterogeneous networks, alternative prompting strategies, and interventions aimed at mitigating bias propagation.

\bibliographystyle{splncs04}
\bibliography{references}

@article{bakshy2015exposure, title={Exposure to ideologically diverse news and opinion on Facebook}, author={Bakshy, Eytan and Messing, Solomon and Adamic, Lada A}, journal={Science}, volume={348}, number={6239}, pages={1130--1132}, year={2015}, publisher={American Association for the Advancement of Science} }

@inproceedings{berjawi2025analyzing, title={Analyzing the Persuasive Strategies of Influencers and News Media on Social Media}, author={Berjawi, Omran and Khatoun, Rida and Fenza, Giuseppe}, booktitle={2025 IEEE/ACS 22nd International Conference on Computer Systems and Applications (AICCSA)}, pages={1--7}, year={2025}, organization={IEEE} }

@inproceedings{berjawi2024multi, title={A Multi-aspect Analysis of Echo Chambers on Video-Sharing Social Media}, author={Berjawi, Omran and Cavaliere, Danilo and Fenza, Giuseppe}, booktitle={International Conference on Advances in Social Networks Analysis and Mining}, pages={197--213}, year={2024}, organization={Springer} }

@inproceedings{borah-mihalcea-2024-towards, title={Towards implicit bias detection and mitigation in multi-agent llm interactions}, author={Borah, Angana and Mihalcea, Rada}, booktitle={Findings of the Association for Computational Linguistics: EMNLP 2024}, pages={9306--9326}, year={2024} }

@inproceedings{chuang-etal-2024-simulating, title={Simulating opinion dynamics with networks of llm-based agents}, author={Chuang, Yun-Shiuan and Goyal, Agam and Harlalka, Nikunj and Suresh, Siddharth and Hawkins, Robert and Yang, Sijia and Shah, Dhavan and Hu, Junjie and Rogers, Timothy}, booktitle={Findings of the association for computational linguistics: NAACL 2024}, pages={3326--3346}, year={2024} }

@inproceedings{cisneros-velarde-2025-biases, title={Biases in opinion dynamics in multi-agent systems of large language models: A case study on funding allocation}, author={Cisneros-Velarde, Pedro}, booktitle={Findings of the Association for Computational Linguistics: NAACL 2025}, pages={1889--1916}, year={2025} }

@article{degroot1974reaching, title={Reaching a consensus}, author={DeGroot, Morris H}, journal={Journal of the American Statistical association}, volume={69}, number={345}, pages={118--121}, year={1974}, publisher={Taylor \& Francis} }

@article{turn0search0, title={Impact of mindset types and social community compositions on opinion dynamics: A large language model-based multi-agent simulation study}, author={Ding, Guozhu and Liu, Zuer and Li, Shan and Cao, Jie and Ye, Zhuohai}, journal={Computers in Human Behavior}, volume={172}, pages={108730}, year={2025}, publisher={Elsevier} }

@article{gallegos2024bias, title={Bias and fairness in large language models: A survey}, author={Gallegos, Isabel O and Rossi, Ryan A and Barrow, Joe and Tanjim, Md Mehrab and Kim, Sungchul and Dernoncourt, Franck and Yu, Tong and Zhang, Ruiyi and Ahmed, Nesreen K}, journal={Computational linguistics}, volume={50}, number={3}, pages={1097--1179}, year={2024} }

@article{hegselmann2015opinion, title={Opinion dynamics and bounded confidence: Models, analysis and simulation}, author={Hegselmann, Rainer and Krause, Ulrich}, journal={Journal of Artificial Societies and Social Simulation}, volume={5}, number={3}, year={2002} }

@inproceedings{turn0academia26, title={Towards simulating social influence dynamics with llm-based multi-agents}, author={Lin, Hsien-Tsung and Huang, Pei-Cing and Ku, Chan-Tung and Hsu, Chan and Shieh, Pei-Xuan and Kang, Yihuang}, booktitle={2025 IEEE International Conference on Information Reuse and Integration and Data Science (IRI)}, pages={307--312}, year={2025}, organization={IEEE} }

@article{turn0search10, title={Homophily-induced emergence of biased structures in llm-based multi-agent ai systems}, author={Mehdizadeh, Aliakbar and Hilbert, Martin}, journal={Social Network Analysis and Mining}, volume={15}, number={1}, pages={1--25}, year={2025}, publisher={Springer} }

@article{moscovici1969influence,
  title={Influence of a consistent minority on the responses of a majority in a color perception task},
  author={Moscovici, Serge and Lage, Elisabeth and Naffrechoux, Martine},
  journal={Sociometry},
  pages={365--380},
  year={1969},
  publisher={JSTOR}
}

@article{piao2025emergence, title={Emergence of human-like polarization among large language model agents}, author={Piao, Jinghua and Lu, Zhihong and Gao, Chen and Xu, Fengli and Hu, Qinghua and Santos, Fernando P and Li, Yong and Evans, James}, journal={arXiv preprint arXiv:2501.05171}, year={2025} }

@article{turn0search6, title={Performance and biases of large language models in public opinion simulation}, author={Qu, Yao and Wang, Jue}, journal={Humanities and Social Sciences Communications}, volume={11}, number={1}, pages={1--13}, year={2024}, publisher={Palgrave} }

@inproceedings{wang-etal-2025-decoding, title={Decoding echo chambers: LLM-powered simulations revealing polarization in social networks}, author={Wang, Chenxi and Liu, Zongfang and Yang, Dequan and Chen, Xiuying}, booktitle={Proceedings of the 31st international conference on computational linguistics}, pages={3913--3923}, year={2025} }

@article{turn0search1, title={Social opinions prediction utilizes fusing dynamics equation with LLM-based agents}, author={Yao, Junchi and Zhang, Hongjie and Ou, Jie and Zuo, Dingyi and Yang, Zheng and Dong, Zhicheng}, journal={Scientific Reports}, volume={15}, number={1}, pages={15472}, year={2025}, publisher={Nature Publishing Group UK London} }

@article{turn0academia20, title={MTOS: A LLM-Driven Multi-topic Opinion Simulation Framework for Exploring Echo Chamber Dynamics}, author={Zuo, Dingyi and Zhang, Hongjie and Ou, Jie and Feng, Chaosheng and Liu, Shuwan}, journal={arXiv preprint arXiv:2510.12423}, year={2025} }

@article{turn0search2, title={Interactive simulation and visual analysis of social media event dynamics with LLM-based multi-agent modeling}, author={Cheng, Zichen and Lin, Ziyue and Yang, Yihang and Wei, Zhongyu and Chen, Siming}, journal={Visual Informatics}, pages={100260}, year={2025}, publisher={Elsevier} }

\end{document}